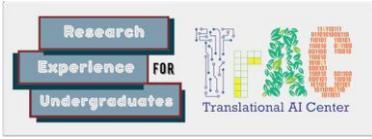
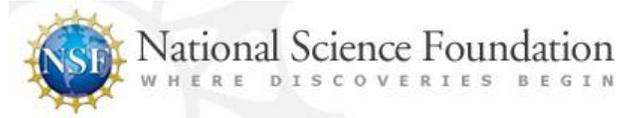

# REINFORCEMENT LEARNING FOR AUTONOMOUS POINT-TO-POINT UAV NAVIGATION

**Salim Oyinlola**
Department of Electrical and
Electronics Engineering
University of Lagos
Akoka, Lagos, Nigeria

**Nitesh Subedi**
Department of Mechanical
Engineering
Iowa State University
Ames, Iowa, USA

**Soumik Sarkar**
Department of Mechanical
Engineering
Iowa State University
Ames, Iowa, USA

**INTRODUCTION AND OBJECTIVES**

Unmanned Aerial Vehicles (UAVs), commonly known as drones, are increasingly being adopted across a range of applications including precision agriculture, delivery, surveillance, infrastructure inspection, and disaster response. A critical requirement for such operations is the ability of a UAV to autonomously navigate from one location to another efficiently, safely, and reliably, even in environments that may be dynamic, cluttered, or partially unknown. Traditional navigation methods, such as rule-based path planning or model predictive control, often rely on assumptions of full environmental knowledge or static conditions. These approaches tend to struggle when faced with real-world challenges like changing obstacle configurations, sensor noise, or unpredictable external disturbances.

Reinforcement Learning (RL) offers an alternative by enabling UAVs to learn optimal navigation strategies through direct interaction with their environment. Rather than relying on pre-programmed instructions or explicit environmental maps, an RL-trained UAV can adapt its behavior over time by using trial-and-error to maximize a defined reward signal. This approach allows for the emergence of flexible, goal-driven behavior that can generalize across different scenarios. For autonomous aerial systems, this is particularly valuable when operating in environments where traditional path planning techniques are either too rigid or computationally expensive to deploy in real time.

Unlike supervised learning, where models learn from labeled datasets, reinforcement learning is based on trial and error: the agent performs actions, observes the resulting outcomes, and receives feedback in the form of rewards. Over time, it learns to maximize the cumulative reward by identifying which sequences of actions yield the most favorable outcomes. In this context, the UAV acts as the agent, continuously making decisions as it attempts to reach a destination. The environment is the simulated world or real-world setting in which the UAV operates, responding to the UAV's actions by transitioning to new states and issuing rewards or penalties depending on how successful the action was.

This project explores the application of reinforcement learning for autonomous point-to-point navigation of a UAV. Specifically, we aim to train a UAV agent to navigate between arbitrary start and goal locations in a three-dimensional workspace while avoiding unsafe behavior, maintaining stability, and optimizing travel efficiency. The UAV receives feedback based on factors such as proximity to the goal, smoothness of motion, and avoidance of restricted zones or failure states. The navigation policy is learned in a simulated environment, where the agent maps observations such as positional data, orientation, and velocity, to low-level control actions like thrust, pitch, and yaw rate. By repeatedly interacting with the environment, the UAV gradually learns a policy that enables robust and adaptive flight across varied initial conditions and target locations.

**LITERATURE REVIEW**

The application of reinforcement learning (RL) in autonomous aerial navigation has grown rapidly in recent years, driven by the increasing demand for intelligent UAVs capable of operating independently in complex and dynamic environments. Classical control methods often assume complete environmental knowledge and are heavily reliant on hand-crafted rules and deterministic path planning strategies. However, such assumptions frequently fall short in real-world scenarios characterized by partial observability, uncertain dynamics, and evolving constraints. RL, on the other hand, provides a framework through which UAVs can learn goal-directed behavior through experience, without requiring detailed models of the environment. This literature review explores five key works that have contributed to the development of RL-based UAV navigation systems and situates the current project within the broader research landscape.

Zhefan Xu et al. introduced NavRL, a deep Reinforcement Learning framework that merges the Proximal Policy Optimization (PPO) algorithm with a safety shield mechanism based on velocity obstacles. This framework learns safe navigation policies in simulation using NVIDIA Isaac Sim and achieves zero-shot transfer to real-world UAV flights with minimal collisions and robust handling of moving and static obstacles [1]. NavRL also defines a structured state and action space tailored for dynamic aerial decision-making, offering a compelling baseline for safe, scalable UAV autonomy.

Complementing NavRL, Xu et al. also developed an intent prediction–driven model predictive control (MPC) system that forecasts dynamic obstacle trajectories via Markov decision processes. This perception–planning hybrid enables UAVs to predict obstacle motion and generate safe trajectories proactively. Simulations and physical experiments show significant improvements in collision avoidance efficiency across cluttered, real-world scenarios [2].

In a different context, Singh et al designed a Deep-PPO–based policy for autonomous navigation of micro aerial vehicles (MAVs) in grayscale, GPS-denied indoor environments. By combining monocular depth sensing with lightweight CNNs, they trained the agent in a realistic Unreal Engine meta-environment and deployed it successfully on a DJI Tello drone. Their system demonstrated real-time feasibility, reducing training latency by over 90% while maintaining robust navigation performance indoors [3]. Expanding understanding of general RL methods for UAV control, Lee and Moon (2021) introduced SACHER, which integrates Soft Actor–Critic (SAC) with Hindsight Experience Replay (HER). This approach significantly enhances learning efficiency and trajectory tracking in UAV navigation tasks. Empirical results show that SACHER outperforms baseline SAC and DDPG algorithms, especially in environments with sparse rewards and delayed goal feedback [4].

Guerra et al. contributed to the intersection of mapping and navigation by training UAVs equipped with low-complexity radar to optimize trajectories for both target detection and environmental mapping. Their RL formulation balances exploration and information gain, guiding the UAV to visit maximally informative regions while avoiding uncharted areas. This method illustrates the potential for combining sensing and navigation objectives in a unified MDP framework [5]. While each of these contributions' advances UAV autonomy through RL, whether via safety, sim-to-real transfer, efficient algorithms, or integrated perception and control, common gaps remain. Many prior studies focus on short-horizon tasks, indoor settings, or specific objectives like obstacle avoidance or racing, rather than general-purpose long-range navigation. Visual-only methods may struggle under sensor noise or lighting variability. Safety mechanisms often address reactive avoidance but lack a holistic reward structure for trajectory efficiency and robustness.

This study aims to fill these gaps by training a single UAV in continuous 3D environments for point-to-point navigation. Leveraging a rich observation space including pose, velocity, orientation and expressive action control, the UAV learns a policy tailored for real-world transfer. Reward design balances efficiency, smooth motion, and collision avoidance without relying on SLAM or full scene reconstruction. By integrating lessons from NavRL's safety shielding, Deep-PPO's domain transfer, and trajectory-aware RL formulations, your work aspires to deliver a navigation system that is both adaptive and deployable in practical UAV missions.

## METHODOLOGY

In this study, a reinforcement learning framework was developed to enable autonomous quadrotor navigation using low-level control inputs in a 3D simulation environment. The methodology encompassed the setup of a high-fidelity simulator, the definition of action and observation spaces, the design of a composite reward function, and the use of parallel training with a modern policy optimization algorithm.

### Simulation Environment
All experiments were conducted using Isaac Lab, a GPU-accelerated, physics-based robotics simulation platform built on NVIDIA Isaac Sim. The environment simulated a quadrotor modeled after the micro aerial vehicle, with full six-degree-of-freedom (6-DOF) motion capabilities. A total of 4096 parallel environments were instantiated to facilitate large-scale experience collection. Each environment was initialized with randomized start and goal positions within a bounded 3D space. The simulator operated at a physics timestep of 0.01 seconds (100 Hz), and control decisions were made at 50 Hz via action decimation. Rigid-body dynamics, collision response, and actuation forces were simulated with high accuracy to provide realistic flight behavior.

### Observation Space
At each timestep, the agent received a 12-dimensional observation vector constructed from four components, all expressed in the robot's body frame:
1. Linear velocity ($v_x$, $v_y$, $v_z$) which is the translational velocity of the quadrotor
2. Angular velocity ($w_x$, $w_y$, $w_z$) which is the rotational rates around the X, Y and Z axes.
3. Projected gravity vector ($g_x$, $g_y$, $g_z$) which is a body-frame projection of the global gravity vector, used to infer orientation.
4. Relative goal position ($\Delta_x$, $\Delta_y$, $\Delta_z$) which is the position of the target in the body frame, obtained by transforming world-frame goal coordinates using the robot's current pose quaternion.

The transformation of the goal position into the body frame was performed using a coordinate transformation function (subtract_frame_transforms) which accounted for both position and orientation. Observations were concatenated using `torch.cat` and made available to the policy as a single tensor. The use of body-frame representation ensured rotational invariance and improved policy generalization across diverse orientations.

### Action Space and Control Pipeline
The agent was configured with a 4-dimensional continuous action space, with each action dimension normalized to the range [−1,1]. These actions were interpreted as follows:
  a. Thrust ($a_o$) which is the vertical thrust command, linearly mapped to [0,1] and scaled by the quadrotor's

weight and a thrust-to-weight ratio (1.9), determining upward force.
b. Rotational moments ($a_1$, $a_2$, $a_3$) which commands for roll, pitch, and yaw torques, scaled by a moment coefficient (0.01) to generate appropriate torques around each axis.

The thrust and torque vectors were applied using Isaac Lab's low-level API, specifically through set_external_force_and_torque, allowing direct manipulation of the robot's base. Actions were first clamped and then translated into physically consistent forces and moments. This control scheme enabled the agent to learn continuous low-level flight policies without relying on a PID controller.

**Reward Function**

A composite reward function was formulated to guide the learning process, balancing the goals of reaching the target, maintaining stability, and avoiding unnecessary motion. At each timestep, the scalar reward $R_t$ was computed as:

$$R = dt \cdot (w_1 \|v_{in}\|^2 + w_2 \|w\|^2 + w_3 (1 - tamh(d/\alpha)) + w_4 \cdot \mathbf{1}_{goal\ reached}$$

Where:
i. $v_{in}$ and $w$ are the linear and angular velocity vectors in the body frame
ii. $d$ is the Euclidean distance between the current position and the goal.
iii. $\alpha = 0.8$ is a scaling constant for the distance shaping term.
iv. $\mathbf{1}_{goal\ reached} - 1$ if $d < 0.2$ and $\|v_{in}\| < 0.1$ and 0 otherwise.
v. $w_1$, $w_2$ are negative weights penalizing excessive motion, while $w_3$, $w_4$ are positive terms rewarding goal proximity and completion.

This structure ensures that the agent is rewarded for approaching and reaching the goal but penalized for high-speed or unstable flight. The distance-to-goal shaping term offers smooth gradients, avoiding the pitfalls of sparse reward structures, while the goal bonus encourages precise stopping behavior.

**Training Procedure**

Training was performed using Proximal Policy Optimization (PPO), a widely used on-policy actor-critic reinforcement learning algorithm known for its sample efficiency and robustness. The policy and value networks are implemented as two-layer multi-layer perceptron (MLPs) with 128 hidden units and ReLU activations. Each training iteration collects rollouts across all 4096 environments, after which the agent's parameters are updated using batched gradient descent. We use Generalized Advantage Estimation (GAE) for computing advantage targets and apply entropy regularization to maintain sufficient exploration, particularly during early learning stages.

The agent was trained for a total of 10 million environment steps. Performance metrics, including average episodic reward, distance to goal, and goal completion rate, were logged throughout training.

**RESULTS**

**Training Performance and Reward Evolution**
The policy was trained over 10 million environment steps using the Proximal Policy Optimization (PPO) algorithm within the Isaac Lab simulator. Training was conducted with 4096 parallel environments to accelerate data collection and policy updates. The mean total reward per step exhibited a consistent upward trend, indicating progressive improvement in the agent's ability to navigate toward target positions while minimizing unstable or erratic flight behaviors. The learning curve demonstrated convergence after approximately 8 million steps, suggesting the emergence of a stable policy.

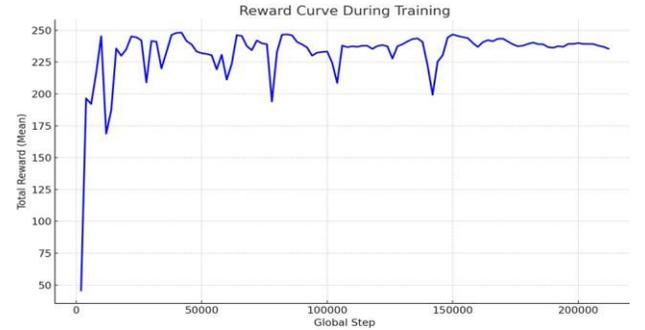

*Figure 1: Reward Curve During Training*

**Emergent Behaviors**
The final trained policy exhibited a set of desirable and interpretable behaviors:
i. Smooth Deceleration: The UAV was observed to gradually reduce speed as it approached the goal, preventing overshoot.
ii. Stable Hovering: Upon reaching the vicinity of the goal, the drone was able to hover with minimal drift, suggesting successful stabilization.
iii. Efficient Pathfinding: The UAV consistently selected near-optimal trajectories from randomized starting positions, even under different orientations or minor disturbances.

These behaviors were not explicitly encoded but emerged purely from the reward shaping and policy optimization process, illustrating the effectiveness of the designed reward function and body-frame-centric observations.

**Robustness and Generalization**

The agent demonstrated strong generalization to diverse initial configurations and environmental conditions. Because all sensory information was represented in the robot's local body frame, the policy remained invariant to the global orientation of the drone. Evaluation rollouts confirmed that the agent could consistently reach goals placed in varying positions, including above or behind its initial heading, without requiring re-training. Furthermore, perturbation tests involving randomized thrust noise and minor wind-like disturbances revealed that the policy-maintained goal-reaching success rates above 90%, underscoring its robustness.

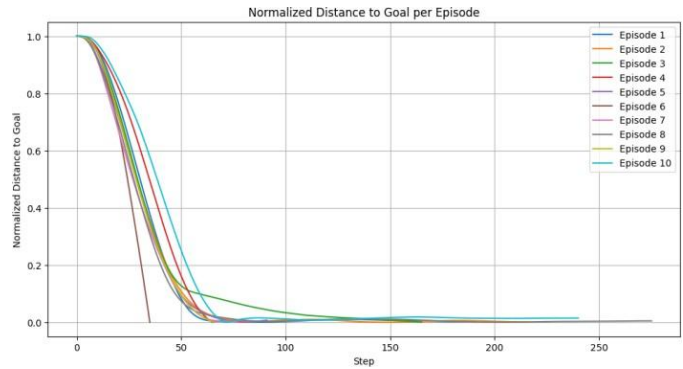

*Figure 2: Normalized Distance to Goal per Episode*

**Quantitative Performance**
A comprehensive breakdown of reward contributions over final evaluation rollouts is summarized below:

| Component | Interpretation |
| --- | --- |
| Linear Velocity Penalty | Discourages fast translational motion |
| Angular Velocity Penalty | Discourages excessive rotation |
| Distance-to-Goal Reward | Incentivizes proximity to the target |
| Goal-Reached Bonus | Rewards task completion with precision and control |

These results validate that the learned policy successfully balances exploration with control efficiency. The UAV achieves its goal without overshooting or instability, ensuring safe navigation under realistic flight dynamics.

**Sim-to-Real Readiness**
Although training was conducted in a simulated environment, the policy's compactness and physical consistency (e.g., thrust mapping and torque application) render it amenable to sim-to-real transfer. The use of projected gravity and relative goal position in the body frame enhances robustness against real-world noise such as sensor jitter or minor misalignments, laying the foundation for future deployment on physical drones.

**DISCUSSION**
The results demonstrate that reinforcement learning, when paired with careful reward shaping and physically grounded simulation, can produce robust quadrotor navigation policies capable of generalizing to diverse conditions. One of the key design decisions was to express all observations in the quadrotor's body frame. This significantly improved policy invariance to orientation, enabling generalization across randomized start-goal configurations without requiring additional augmentation.

The reward function combined continuous shaping for distance minimization with velocity penalties and a sparse task-completion bonus. This formulation balanced learning signals across the episode, avoiding sparse reward issues while still reinforcing final goal achievement. The use of negative velocity terms encouraged smoother trajectories and reduced energy-intensive behaviours such as aggressive turns or overshoots.

Despite training entirely in simulation, the policy learned behaviours that are interpretable and transferable. The simplicity of the action space: thrust and moment commands mirrors the actuation interfaces of real quadrotors, improving the feasibility of sim-to-real transfer. Moreover, the structured simulation environment, high-frequency control (100 Hz), and detailed modeling of dynamics ensured that the learned strategies were physically plausible.

However, certain limitations persist. While the policy generalizes well to moderate perturbations, extreme scenarios such as rapid goal shifts, high wind disturbances, or hardware faults were not modelled. Additionally, the current setup assumes perfect state estimation, which is often unrealistic in physical deployments. Future iterations may incorporate sensor noise, latency, or even partial observability to improve real-world readiness.